\documentclass{article}
\usepackage{spconf,amsmath,epsfig}
\usepackage{subfig}

\newcommand{\etal}{et al. }
\newcommand{\eg}{e.g. }

\begin{document}\sloppy

\title{Diversity in Object Proposals}
%
\name{Anton Winschel \hspace{2cm} Rainer Lienhart \hspace{2cm} Christian Eggert}
\address{{\normalsize Multimedia Computing and Computer Vision Lab} \\
                {\normalsize University of Augsburg, Germany} \\
               {\normalsize \{anton.winschel, rainer.lienhart, christian.eggert\}@informatik.uni-augsburg.de}}

\maketitle



\begin{abstract}
Current top performing object recognition systems build on object proposals as a preprocessing step. Object proposal algorithms are designed to generate candidate regions for generic objects, yet current approaches are limited in capturing the vast variety of object characteristics. In this paper we analyze the error modes of the state-of-the-art Selective Search object proposal algorithm and suggest extensions to broaden its feature diversity in order to mitigate its error modes. We devise an edge grouping algorithm for handling objects without clear boundaries. To further enhance diversity, we incorporate the Edge Boxes proposal algorithm, which is based on fundamentally different principles than Selective Search. The combination of segmentations and edges provides rich image information and feature diversity which is essential for obtaining high quality object proposals for generic objects. For a preset amount of object proposals we achieve considerably better results by using our combination of different strategies than using any single strategy alone.
\end{abstract}

\begin{keywords}
Object Proposals, Selective Search, Edge Boxes
\end{keywords}


\section{Introduction}

\begin{figure}[h]
\begin{center}
\includegraphics[width=\linewidth]{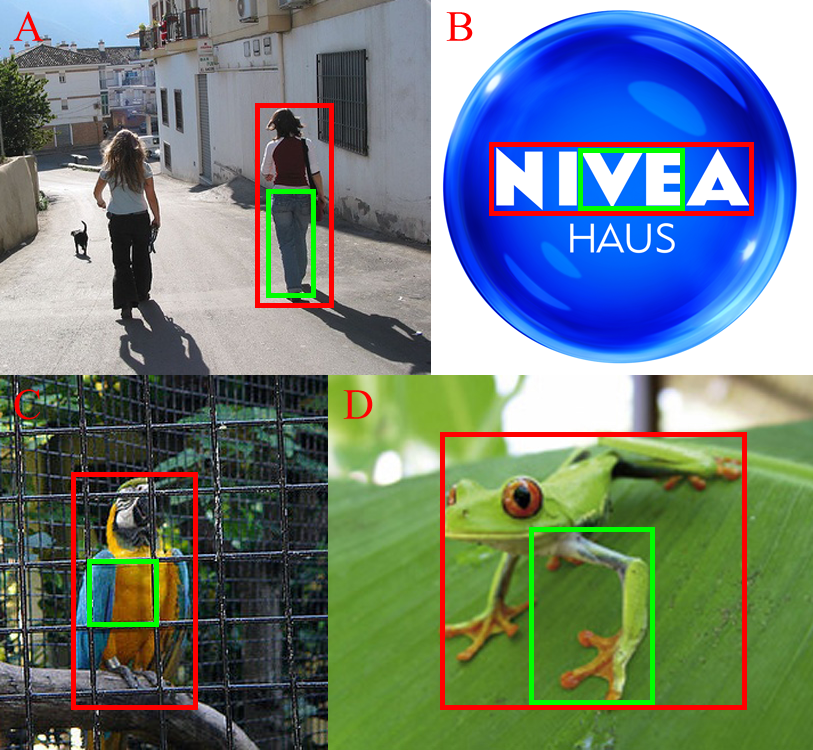}
\caption{Examples of Selective Search (SS) error modes. From left to right and top to bottom: error modes (A), (B), (C) and (D). Red boxes denote ground truth annotations and green boxes denote the best out of 2000 SS proposals.}
\label{error_modes}
\end{center}
\end{figure}

Recently established object recognition pipelines (\cite{Wang14, Girshick14, Szegedy14}) use sophisticated object proposal generators to select potential object locations which are subsequently passed to a detector. This preprocessing step replaces the expensive exhaustive search paradigm which classifies many unnecessary image regions. By means of object proposals only a tiny fraction of all possible locations have to be examined leaving more processing time for exploiting more complex and computationally expensive classifiers. However, object proposals come with a significant drawback: objects that are excluded within this preprocessing step have no chance anymore to be found. For that reason object proposal algorithms aim to satisfy the following properties: (1) their hypotheses must be accurate in localization and carefully chosen as the number of proposals is limited, (2) a high recall should be achieved in order to localize every object and (3) their computation should be efficient in order to minimize the overhead in the recognition pipeline.

Satisfying properties (1), (2) and (3) adequately is very challenging as generic objects can vary greatly in appearance. Current proposal algorithms are not able to handle the variety of object characteristics. Typically a proposal algorithm is based on a single image representation, \eg segmentations or edges, constraining the available image information in advance. In this paper we explore the impact of combining fundamentally different strategies.

First we analyze the error modes of Selective Search  (SS) \cite{Uijlings13}, a state-of-the-art proposal algorithm. We work out a weakness of SS in localizing objects without well-defined closed boundaries. Our edge grouping algorithm VH-Connect is designed to resolve this problem. We also explore the impact of diversification by combining SS that is based on segmentations with an edge-based proposal approach, namely Edge Boxes (EB) \cite{Zitnick14}.

By combining fundamentally different strategies we enhance the proposal diversity and cover more object characteristics. In our experiments we show that the combined strategies perform better than any single strategy alone.

Our contributions are as follows: we (1) present an analysis of the error modes of SS, (2) propose the edge grouping algorithm VH-Connect which is used to localize objects consisting of multiple but not necessarily connected components, and (3)  show that the combination of conceptually different proposal algorithms achieves better results than any single strategy with the same amount of object proposals.


\section{Related work}

Object proposal methods aim at generating object locations for any potential object class. For this purpose it is necessary to clarify the distinction of generic objects. Alexe \etal \cite{Alexe10} argue that generic objects cannot be designated by a single property. They suggest that any conceivable object satisfies at least one of the following attributes: (1) a clear boundary, (2) a different appearance from its surroundings, and (3) saliency. We consider this definition applicable and employ it for our understanding of generic objects.

In a profound analysis Hosang \etal \cite{Hosang15} compare common object proposal methods. They discover a strong correlation between recall at high IoU thresholds and detection performance. Thus for proposal generators it is vital to achieve high recall and accurate localization.

Zhu \etal \cite{Zhu15} analyze the impact of object characteristics on common proposal methods. They observe that object appearance influences the performance of proposal methods. Objects with homogeneous foreground or background are more likely to be proposed than objects containing structures. This could be an indication of a lack of feature diversity in common proposal methods.

Chen \etal \cite{Chen15} describe a way of improving object proposals by using box refinement. They use segmentation in order to align boxes and to improve localization accuracy. Our approach for improving proposal performance is to analyze and resolve weaknesses of Selective Search. We notice a difficulty of Selective Search in generating good hypotheses for some object characteristics, hence box refinement has little effect in these cases.

Selective Search \cite{Uijlings13} is a widely used proposal generator. It is a greedy algorithm that hierarchically combines segments from initial oversegmentations. Due to its state-of-the-art performance we analyze its error modes and propose a way to resolve them.

The Edge Boxes proposal algorithm \cite{Zitnick14} also yields state-of-the-art results. It starts by computing an edge-map of the input image. Then it uses a sliding window approach to generate multiple proposals, which are sorted according to an objectness value that is derived from the edge distribution.

SS and EB are both considered to be amongst current top performing proposal generators (see \cite{Hosang15, Zhu15}). On grounds of their conceptually different methodologies we analyze the impact of diversity in object proposals.


\section{Error modes of Selective Search}

The basis of Selective Search \cite{Uijlings13} is a pixelwise oversegmentation of an image. Afterwards, a similarity value is computed between all neighboring segments. The similarity metric is based on color and texture information and the size and relative location of the segments. Subsequently the two most similar neighbors are merged to a single segment. Each generated segment with its enclosing bounding box constitutes an object hypothesis.

SS is widely used in top-performing object recognition frameworks like \cite{Girshick15, Girshick14, Wang14}. It is fast in computation, compared to equally well performing proposal methods, and generates accurate object locations. However there are certain cases where SS performs rather poorly or even has conceptual problems. In this section we analyze the four most frequent and most severe error modes of SS.\medskip

{\noindent \bf (A) Neighboring connected components.}
The most common error mode occurs with objects consisting of multiple spatially connected components with different color or texture. In this configuration, single components usually are correctly constructed during the merging steps (\eg pants as a single component and shirt as another component). In order to generate a good object hypothesis, the individual object parts should merge to a single region. But the probability of merging the object parts together is often roughly the same as merging an object part with a neighboring background region. Once the background is merged with an object part, it is unlikely that a good hypothesis will be generated. A typical example is shown in fig. \ref{error_modes}a where a person is wearing trousers that are different in color than the upper body clothes. In the course of the merging steps the lower body is already merged with the background before it can be combined with the upper body.\medskip

{\noindent \bf (B) Spatially divided components.}
A similar case occurs with objects consisting of multiple spatially divided components. In this particular case there are background segments between the segments of an object such that the object has no closed boundary. However, SS is restricted to merge only directly neighboring segments. Hence a background segment must merge with an object segment before the multiple object segments can be merged, making an enclosing object hypothesis impossible in many cases. An example is shown in fig. \ref{error_modes}b. Here the company logo consists of multiple components with segments in between. The entire logo cannot be proposed as a candidate region because of the hierarchical nature of SS. This is a conceptual problem that omits objects without a well-defined boundary.\medskip

{\noindent \bf (C) Partial occlusion.}
Another difficulty is generating good proposals for partially occluded objects. If the object of interest is occluded such that it is divided into multiple parts, SS can hardly generate a tight enclosing object proposal. Even a single continuous line across an object can cause trouble. The merging procedure has to insert a segment from the occluding object into the object region set. At this point further merging with the wrong object is encouraged. In fig. \ref{error_modes}c the bird is partially covered by a lattice. SS does not generate an enclosing bounding box because a bird segment eventually merges with a lattice segment.

{\noindent \bf (D) Similarity between background and object.}
The last error mode in our analysis occurs for objects with an obvious contour but with similar color or texture as its surroundings. In this case the similarity measure computed within SS yields similar values between object-object and object-background segments. A merging of object segments and background segments is once more fairly probable. Once this happens, segment merging is more likely to continue to spread into the background. Fig. \ref{error_modes}d shows a frog that is clearly distinguishable from the background, yet in early iterations some object segments get merged with the background creating poor object proposals.

{\noindent \bf Discussion.}
In general the identified errors of SS are based on the hierarchical merging and the negligence of the holistic composition of the image. The greedy principle of merging neighboring segments with a maximum similarity may lead in some cases to disadvantageous merging of object segments with background. In order to counteract irreversible false merges, SS uses a simple diversification strategy. In its fast version, the algorithm is re-run eight times with different parameter settings. This improves the performance (see \cite{Uijlings13} for details), but the fundamental problems of the algorithm are not solved in this manner. We explore proposal diversification on a higher level by augmenting the proposal strategy. In order to cover the error modes (A) and (B), segment boundaries have to be grouped. As to that, SS is restricted to merge neighboring segments only. This concerns case (B) in particular. In order to resolve error modes (C) and (D), high-level edges have to be incorporated. However, SS is merely provided with segmentations, lacking other important information (\eg edges) to localize any generic object class.

For the purpose of enhancing the diversity in object proposals we combine SS with two conceptually different approaches. The additional algorithms are chosen based on our analysis of SS error modes. We propose the edge grouping algorithm VH-Connect that targets at handling error modes (A) and (B). Grouping neighboring edges enables leaping over edge boundaries which is necessary for generating proposals for multiple component objects. Besides that, we use the efficient Edge Boxes algorithm to address error modes (C) and (D). EB generates proposals based on edge information. We have observed that the partial occlusions and background similarity error modes are better handled using edge-based approaches.


\section{VH-Connect Algorithm}

\begin{figure*}[htp]
\begin{center}
\includegraphics[width=\linewidth]{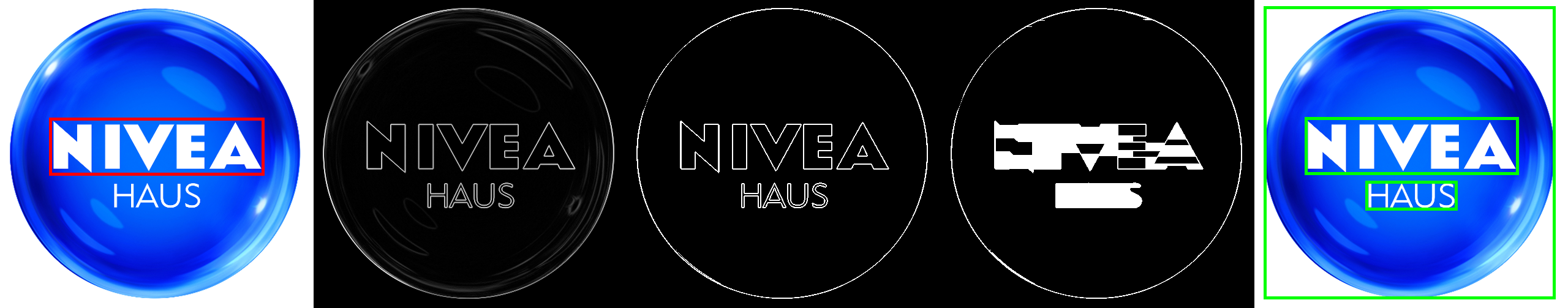}
\caption{VH-Connect, from left to right: (1) original image with ground truth box, (2) morphological gradient, (3) otsu thresholding, (4) a single horizontal grouping step $C_{s,l,k}$ and (5) the generated proposals.}
\label{vh_connect}
\end{center}
\end{figure*}

Objects without closed boundaries and objects consisting of multiple components cause conceptual problems to SS due to its merging methodology. An example is the NIVEA object in fig. \ref{error_modes}b. It lies within a blue background region without a coherent closed boundary. Our VH-Connect algorithm (VH) is designed to generate a few object proposals that cover those objects which are missed by SS due to the error modes (A) and (B). It resolves this particular problem by generating smooth and continuous edges using \cite{Han14} first, and then aggregating vertical and horizontal edges into single structures to establish additional object proposals.\medskip

{\noindent \bf Description of VH-Connect.}
First, a morphological gradient with an elliptical structuring element $b$ is applied to the  grayscale image $f$ of input image $I$ (see \cite{Han14} for details). It is computed as the difference between the  dilated $\oplus$ and eroded $\ominus$ grayscale images as follows:
\begin{equation}
\begin{aligned}
G(x,y) = [f \oplus b](x,y) - [f \ominus b](x,y) = \\
\max_{(s,t) \in b} \{ f(x+s,y+t) \} - \min_{(s,t) \in b} \{ f(x+s,y+t) \}
\end{aligned}
\end{equation}

We use morphological gradients to generate smooth and continuous edges and to prevent broken lines and artifacts. Moreover, the resulting edges are generated without any specified orientation leading to consistent edge thickness (see fig. \ref{vh_connect}(2)). Next, $G(x,y)$ is binarized by Otsu thresholding \cite{Otsu79} to obtain clear edges that define object part contours (see fig. \ref{vh_connect}(3)). Otsu's method computes the optimal threshold that maximizes inter-class variance. 

After this step, edges are grouped by morphological closing operations (= dilation followed by erosion) with one-dimensional structuring elements to form single objects, thus handling cases (A) and (B). In detail, let $S$ be the set of image scales and $L$ the set of kernel lengths. A scale pyramid $I_s$ of images at scales $s \in S$ and multiple horizontal and vertical kernels $h_l = [l \times 1]$ and $v_l = [1 \times l]$ with various lengths $l \in L$ are used to catch different object sizes. The morphological closing operation $\bullet$ is conducted with all combinations of the cartesian product $S \times L$ with $k \in \{h_l, v_l\}$:
\begin{equation}
\begin{aligned}
C_{s,l,k} = [I_s \bullet k_l](x,y) = [(I_s \oplus k_l) \ominus k_l](x,y)
\end{aligned}
\end{equation}

The results are connected structures $s$ whose enclosing boxes generate the VH object proposals (see fig. \ref{vh_connect}(4)  and (5)). In a final post-processing step we filter out the relevant structures by using the objectness-like measure $area(s)~>~p*area(box(s))$, where $p$ denotes the required percentage of the structure's area with respect to the area of the minimal bounding box. \medskip

{\noindent \bf Discussion.}
On average VH generates 200-400 candidate boxes depending on the parameter settings. The main goal is to resolve the error modes (A) and (B) of SS which is achieved by grouping distinct edges into a single structure. For our experiments we use $b$~=~[3$\times$3], $S$~=~\{1,$\frac{1}{2}$,$\frac{1}{4}$\}, and $L$~=~\{9, 15, 30, 45\} on pre-scaled input images with a maximum side length of 1024 px. Relating to SS the algorithm is computationally negligible as morphing operations can be computed very efficiently (see table \ref{time} for processing time measurements).


\section{Combining the Strengths}
It is a significant weakness of current proposal algorithms to exclusively operate on a single image representation. SS is limited to segmentation maps, therefore disregarding some edge details. As a consequence, SS has difficulties to capture the higher level semantic of object contours causing its weakness of handling the cases (C) and (D). The VH-Connect proposal generator does not solve it either. Thus, in order to incorporate edges to our proposal generation pipeline and to enhance the diversity, we utilize an effective edge-based proposal algorithm, namely Edge Boxes \cite{Zitnick14}. Individually, SS and EB yield comparable recall results in many cases (see \cite{Hosang15} for details). However, we have observed that they generate somewhat different object hypotheses as both methods are based on different principles. This in fact benefits the diversification of the proposals.

Instead of using a fixed number of proposals from one particular method, we scatter the requested quantity amongst the three different methods. It is essential that the three methods are carefully picked to conceptually complement each other. In the experiments, we show the gain achieved by a richer image representation in generating generic object proposals.


\section{Experiments}
Our goal is to improve the quality of object proposals measured at a given budget number of proposals by using a combination of conceptually different object proposal generation strategies. In our experiments we analyze the impact of our diversification strategy. \medskip

{\noindent \bf Datasets.} We conduct our experiments on four datasets: (1) the COCO validation set \cite{Lin14} consisting of $\sim$40,500 images from 80 categories, (2) the ImageNet 2013 validation set \cite{Deng09} with $\sim$20,000 images from 200 categories, (3)  the BelgaLogos dataset \cite{Joly09} with $\sim$10,000 images from 37 classes and (4) the FlickrLogos set \cite{Romberg11} with $\sim$8,000 images from 32 logo classes.\medskip

{\noindent \bf Evaluation metrics.} For object proposal evaluation we use common evaluation protocols. Let $c \in C$ denote a class $c$ from the set of all classes $C$ and $G^c$ the set of ground truth annotations of this class in all images; let $L$ be the set of all generated object proposals for all images. The Average Best Overlap (ABO) score averages the maximum intersection over union (IoU) overlap of $L$ with each ground truth annotation $g \in G^c$. The Mean Average Best Overlap (MABO) is calculated by averaging all class ABO values:
\begin{equation}
\begin{aligned}
MABO = \frac{1}{|C|}\sum_{c \in C}[\frac{1}{|G^c|}\sum_{g \in G^c}\max_{l \in L}\mathrm{IoU}(g,l)]
\end{aligned}
\end{equation}

For the {\it recall at IoU threshold} protocol we count all maximum IoU overlaps with a score greater than a specified threshold $t \in [0.5, 1]$:
\begin{equation}
\begin{aligned}
recall(t) = \frac{1}{|G|}\sum_{g \in G}[\max_{l \in L}\mathrm{IoU}(g,l) \geq t]
\end{aligned}
\end{equation}
with $G= \cup G^c$ being the set of all ground truth annotations over all classes.
For the {\it average recall} we compute the area under the {\it recall at IoU threshold} curve for a given number of object proposals:
\begin{equation}
\begin{aligned}
AR = \int_{0.5}^{1} recall(t)\,\mathrm{d}t
\end{aligned}
\end{equation}

{\noindent \bf Experimental setup.}
In our approach we use a combination of SS, VH and EB to enhance the diversity of object proposals. We split the total budget number of proposals amongst all three methods in order to ensure a fair setting. We evaluate the following combinations of object proposals: (1) EB, (2) SS, (3) 50\% SS and 50\% EB, (4) 90\% SS and 10\% VH and (5) 50\% SS, 40\% EB and 10\% VH. Variation in the combination ratios has little influence on the performance, as the proposals of each method are sorted. We conduct time measurements to compare the computational overhead of each method (see table \ref{time}). The VH proposals are generated very efficiently causing an neglectable computational overhead as compared to SS and EB.

\begin{table}[]
\centering
\begin{tabular}{|ll|} 
\hline
Selective Search & 0.9282 [s/image]\\ 
Edge Boxes & 0.2409 [s/image]\\ 
VH-Connect &  0.0276 [s/image]\\ \hline
\end{tabular}
\caption{Average processing time (in seconds) on a single Intel i7 2.9 GHz CPU core for the ImageNet 2013 dataset with an average image size of 482$\times$415 px.}
\label{time}
\end{table}

{\noindent \bf Results.}
At first we study the MABO results using a fixed number of 2000 generated object proposals per experiment (see table \ref{mabo}). Recall that the settings (3), (4) and (5) reduce the number of hypotheses of each strategy to achieve a total number of 2000 proposals. On all datasets SS performs better than EB, especially on BelgaLogos with an performance advantage of 12\% as SS is better at identifying small objects (see table \ref{avgStat}). On Average, settings (3) and (4) give an equal boost on the performance. The combination of SS, VH and EB further increases the MABO performance on all datasets.

As to the {\it recall at IoU threshold} we also use 2000 generated object proposals per experiment. For IoU thresholds between 0.6 and 0.8 EB achieves higher recall scores than SS, except on BelgaLogos where EB performs rather poorly at low thresholds. At high IoU thresholds SS outperforms EB on every dataset. The combination (4) behaves parallel to SS, but at higher recall. Setting (3) outperforms setting (4) on COCO and ImageNet. Eventually, setting (5) outperforms all previous settings on all IoU thresholds.

In our {\it average recall} experiments we vary the number of object hypotheses from 100 to 3000. EB performs better than SS on ImageNet and FlickrLogos. By combining SS, VH and EB we again achieve the best results.

{\noindent \bf Discussion.}
Due to the rigid sampling of EB's sliding windows EB has a poor localization performance denoted by recall at high IoU thresholds. Furthermore EB has difficulties with small objects, as indicated by the evaluation on BelgaLogos. When it comes to smaller IoU thresholds EB outperforms SS. This is mostly due to EB's more consistent covering of the search space by sliding window sampling. SS is similarly biased against large object instances due to its method to rank box proposals according to the merging hierarchy. VH is a crucial improvement when it comes to smaller objects and error modes (A) and (B). Error mode (B) includes many text-like objects that can mostly be found in the logo datasets FlickrLogos and BelgaLogos. 

\begin{table}[]
\centering
\resizebox{0.48\textwidth}{!}{%
\begin{tabular}{|l|llll|} 
\hline
     & COCO  & ImageNet & FlickrLogos &  BelgaLogos \\ \hline
Avg. width  & 104.46 px & 152.29 px & 144.34 px  & 44.26 px \\ 
Avg. height  & 108.04 px & 162.21 px & 103.29 px & 32.46 px \\ 
Avg. ratio & 0.0837 & 0.1688 & 0.0455 & 0.0038 \\ \hline
\end{tabular}
}
\caption{Average width and height of the ground truth annotations and average ratio of the ground truth area with respect to the image size.}
\label{avgStat}
\end{table}

\begin{table}[]
\centering
\resizebox{0.48\textwidth}{!}{%
\begin{tabular}{|l|llll|} 
\hline
     & COCO  & ImageNet &  FlickrLogos & BelgaLogos \\ \hline
EB  & 0.65 & 0.74 & 0.72  & 0.52 \\ 
SS  & 0.70 & 0.76 & 0.75 & 0.64  \\ 
SS+EB  & 0.72 & 0.78 & 0.77 & 0.65  \\ 
SS+VH  & 0.72 & 0.77 & 0.78 & 0.68  \\ 
SS+VH+EB & 0.75 & 0.80 &  0.81 & 0.69  \\ \hline
\end{tabular}
}
\caption{Mean Average Best Overlap (MABO) results using 2000 object proposals.}
\label{mabo}
\end{table}

\begin{figure*}[t]
\captionsetup[subfigure]{labelformat=empty}
\centering
\subfloat[]{\includegraphics[width=0.245\textwidth]{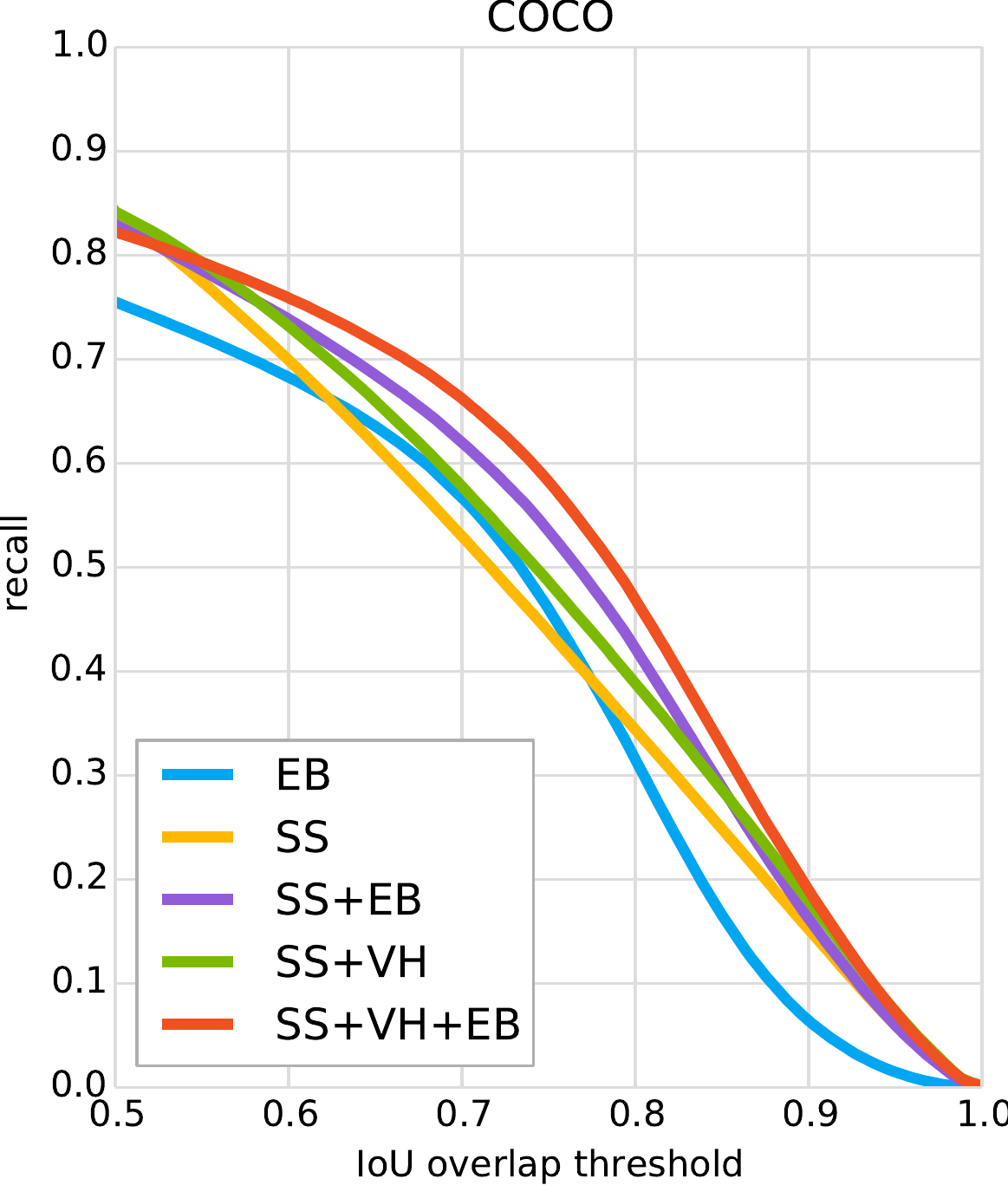}}\vspace*{-1em}\hspace*{0em}
\subfloat[]{\includegraphics[width=0.245\textwidth]{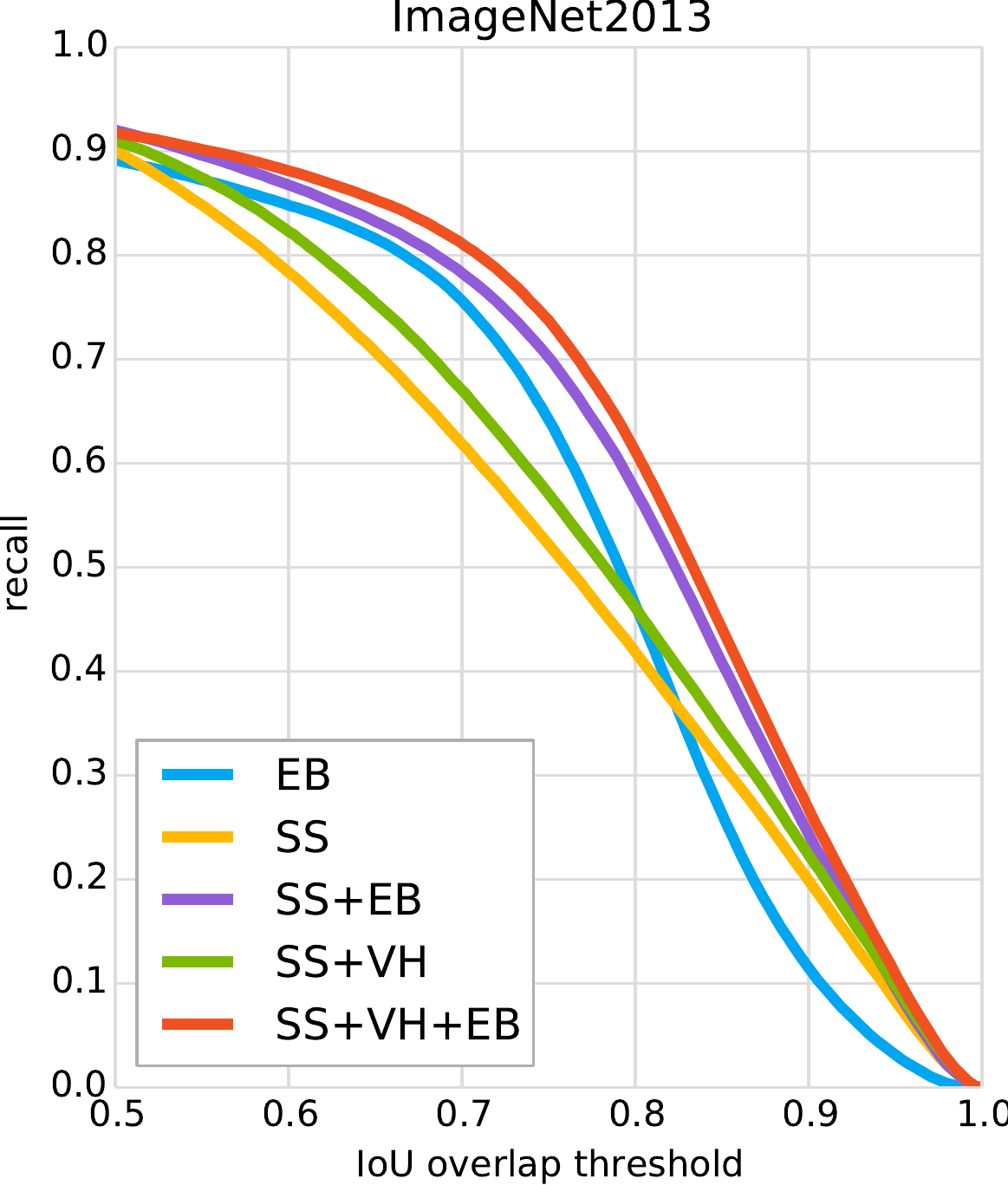}}\hspace*{0em}
\subfloat[]{\includegraphics[width=0.245\textwidth]{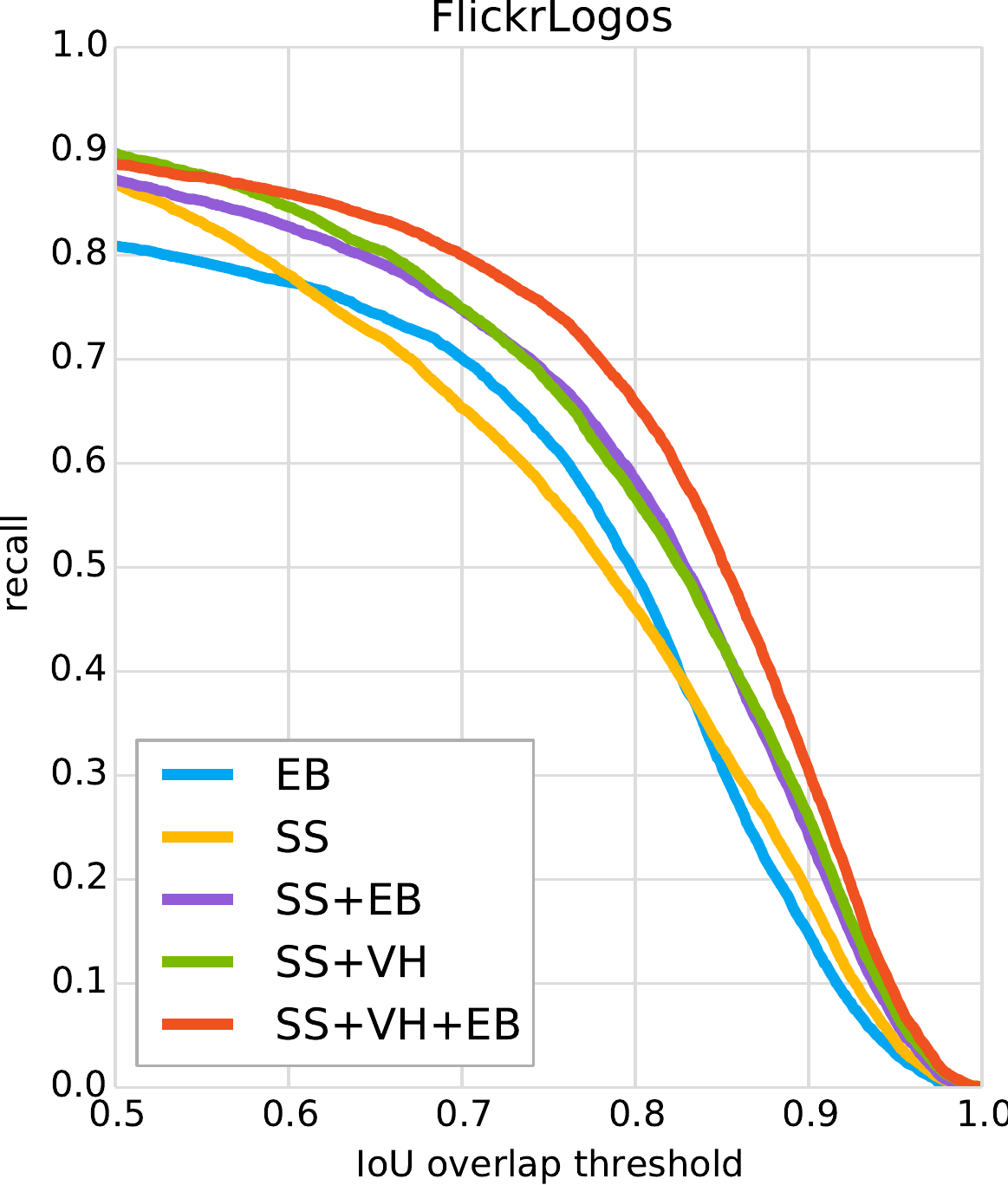}}\hspace*{0em}
\subfloat[]{\includegraphics[width=0.245\textwidth]{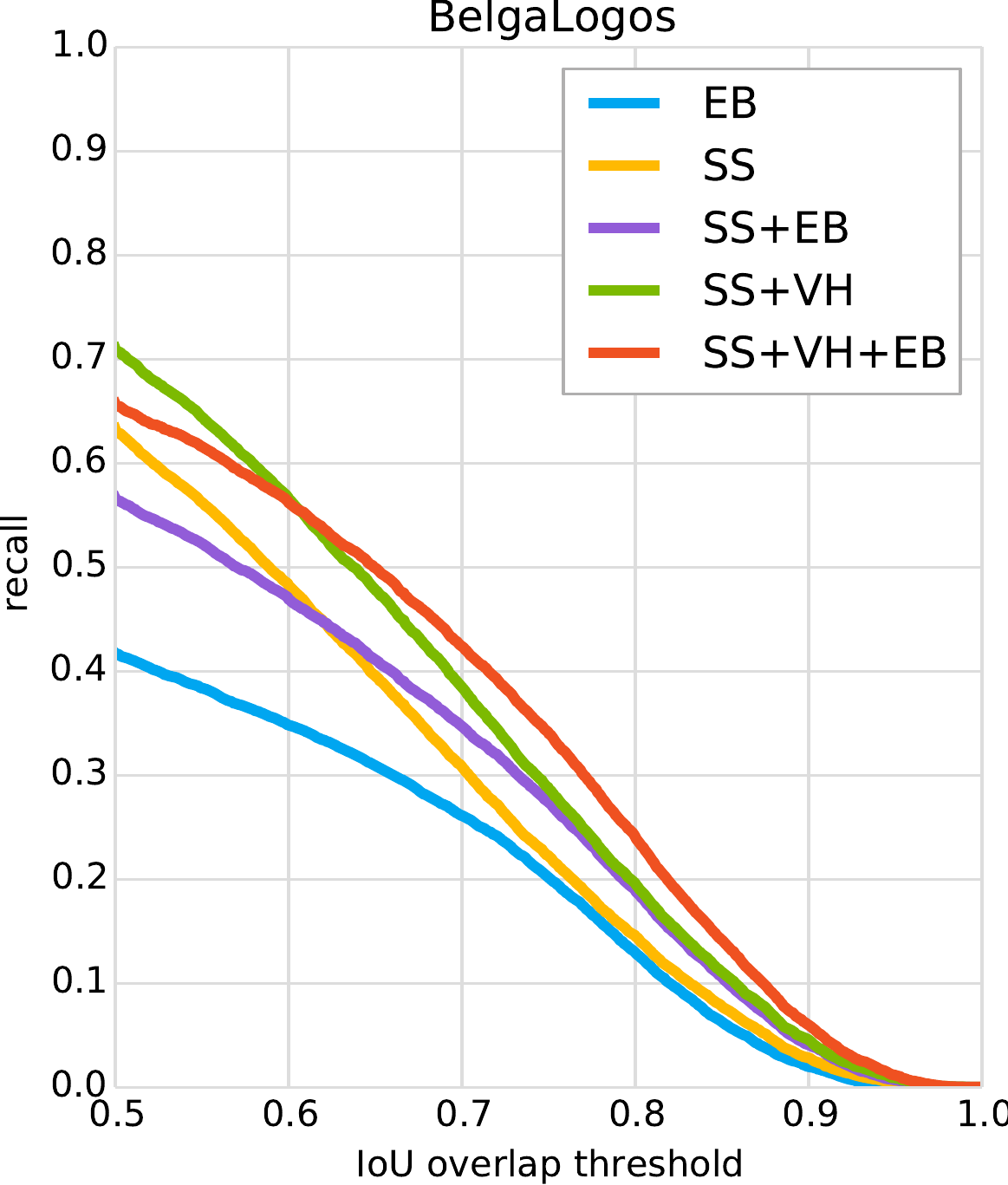}}\\
\subfloat[]{\includegraphics[width=0.245\textwidth]{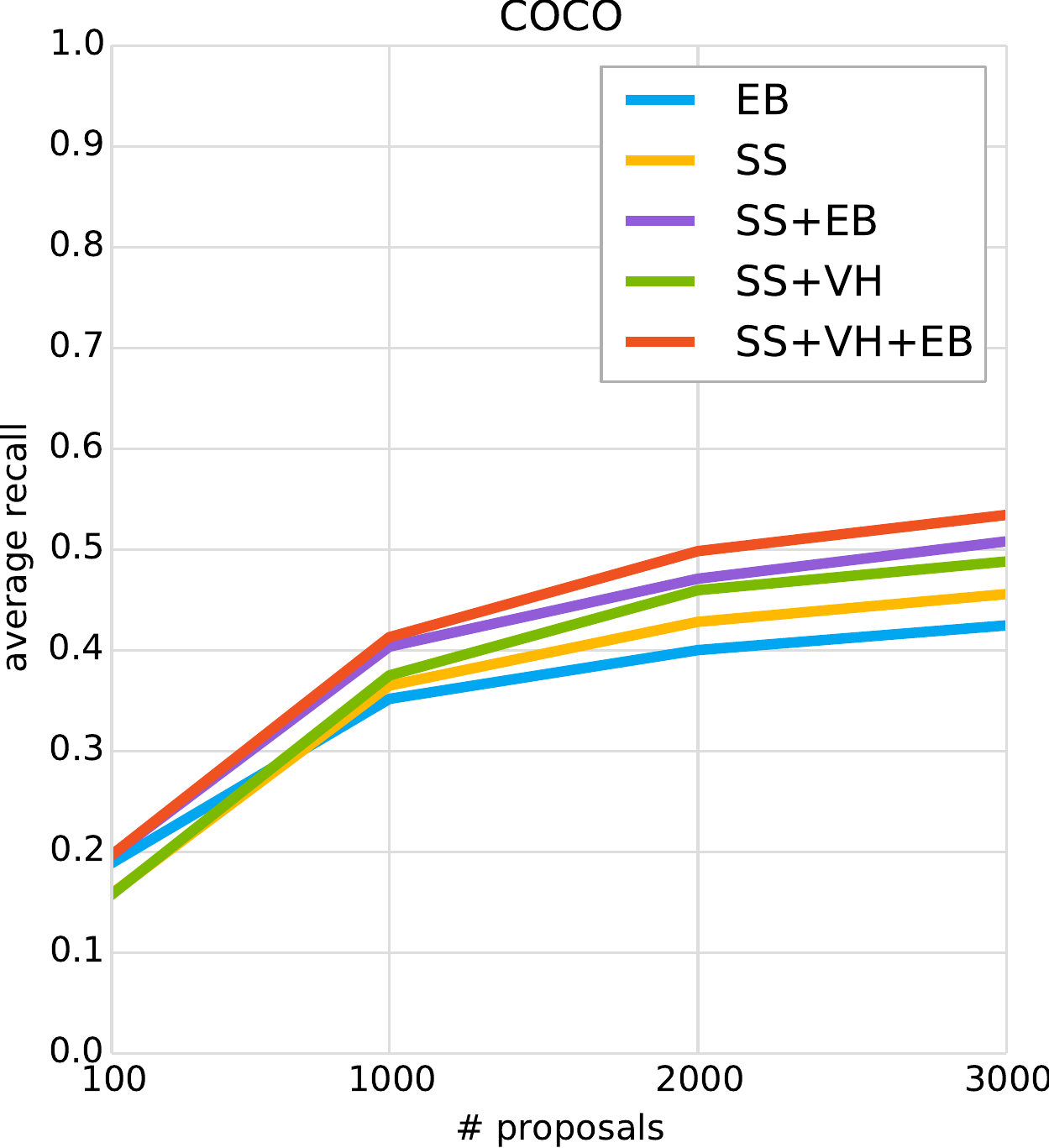}}\hspace*{0em}
\subfloat[]{\includegraphics[width=0.245\textwidth]{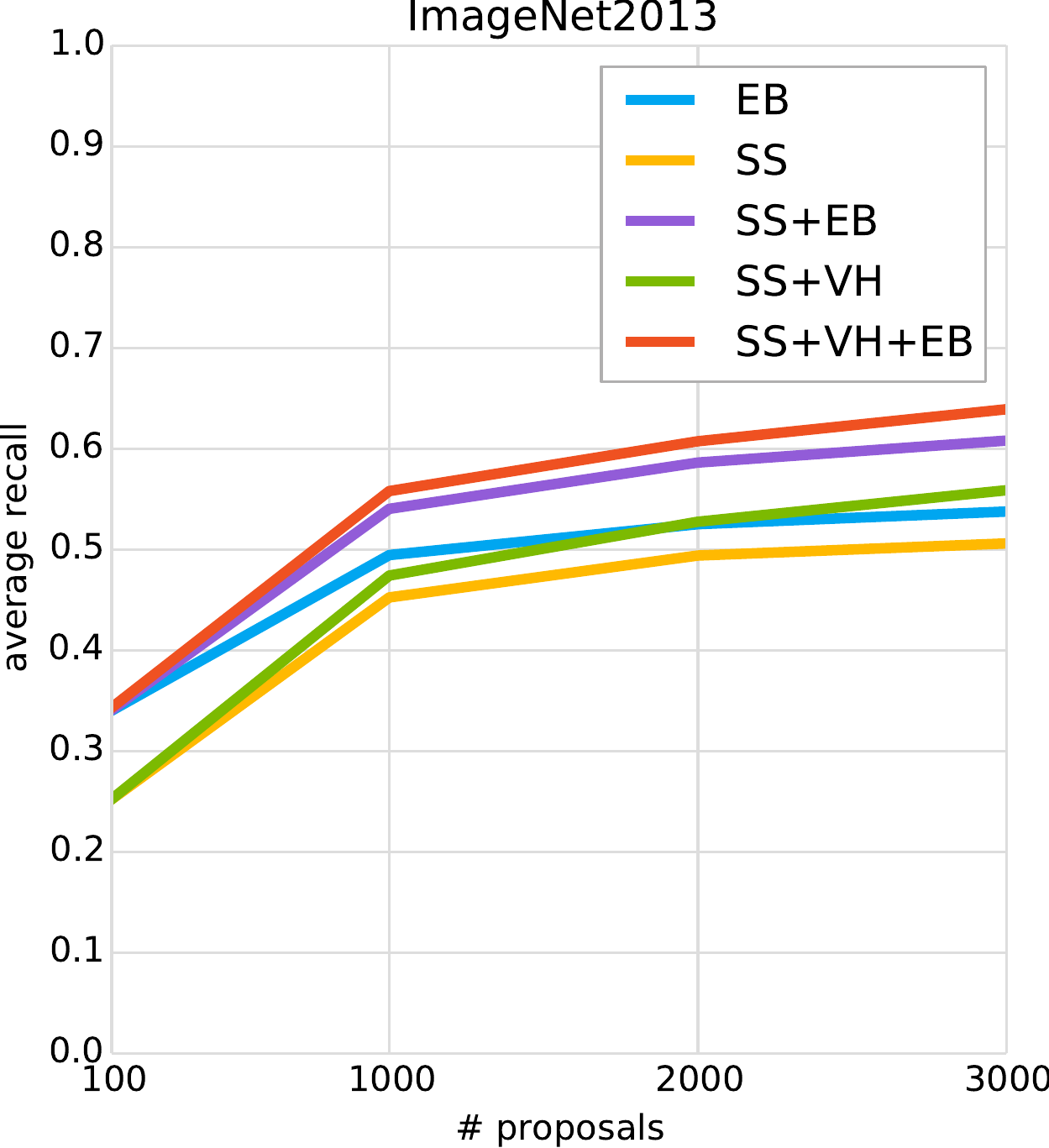}}\hspace*{0em}
\subfloat[]{\includegraphics[width=0.245\textwidth]{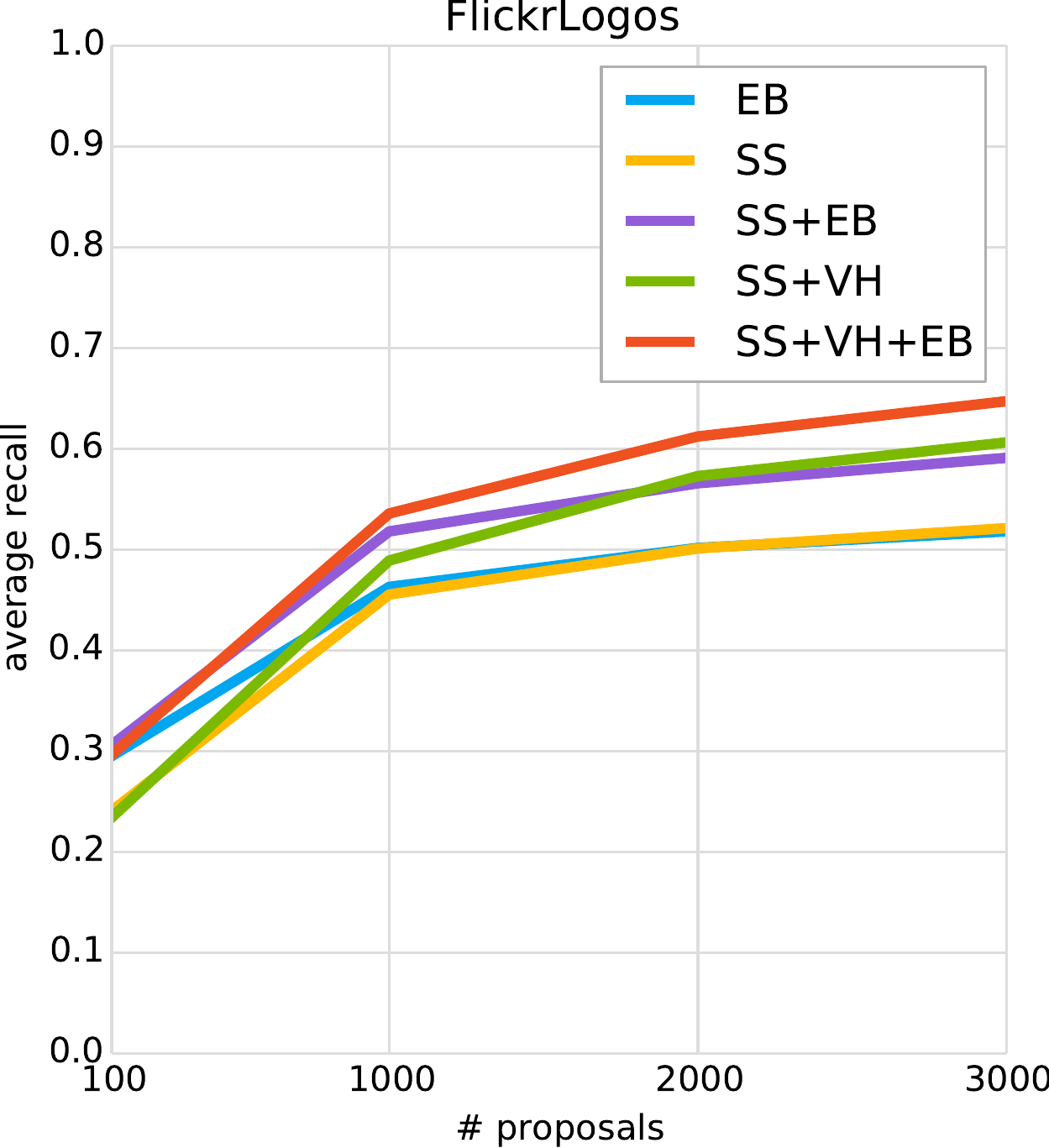}}\hspace*{0em}
\subfloat[]{\includegraphics[width=0.245\textwidth]{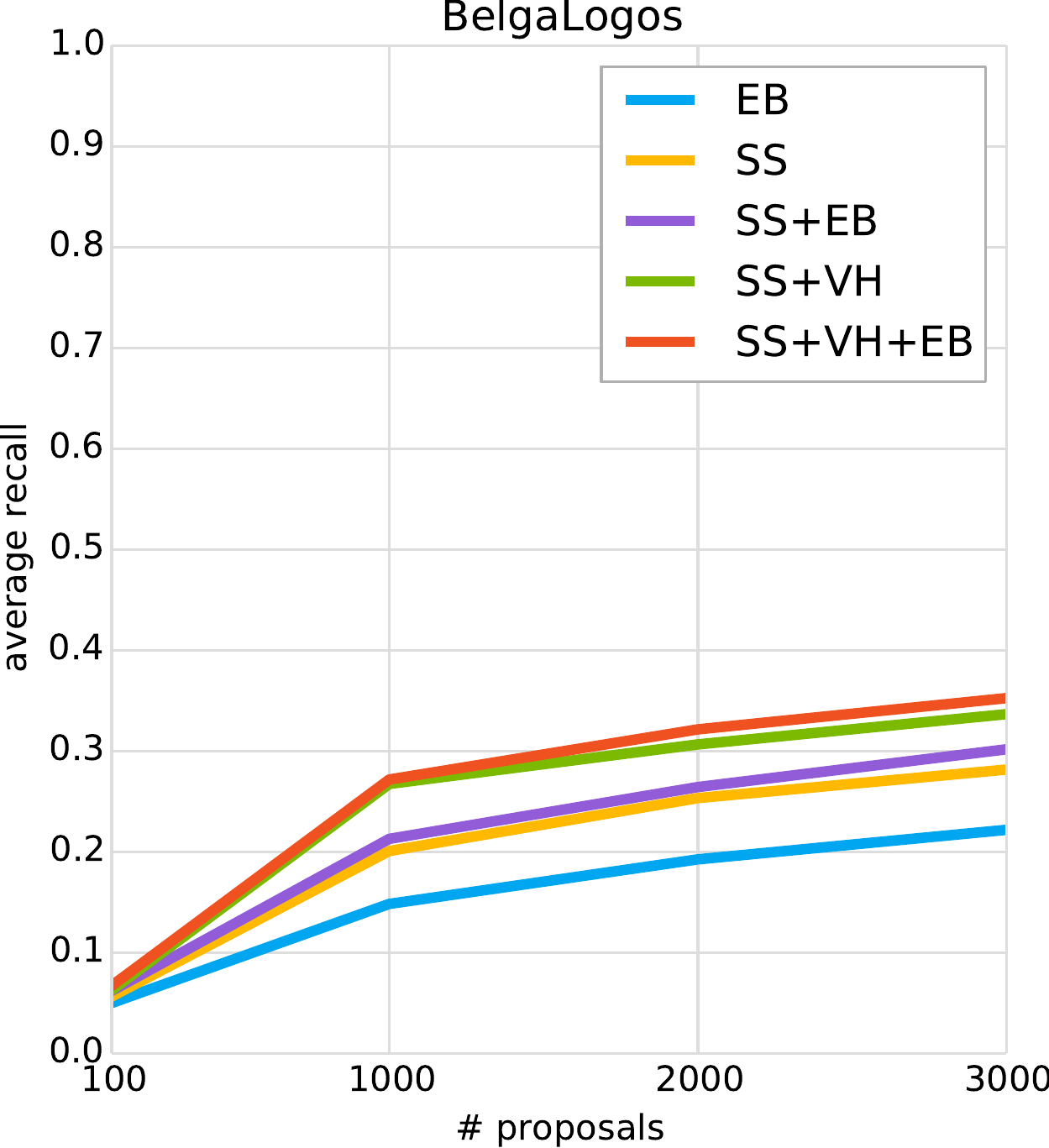}}
\caption{Upper row: recall plotted against IuO threshold at 2000 proposals in total. Lower row: average recall plotted against number of proposals.}
\label{recallFig}
\end{figure*}


\section{Conclusion}
In this paper we have shown that diversity in object proposal generation can improve proposal performance. Generic object classes have many peculiarities that cannot be captured using a single strategy. We have selected the current state-of-the-art Selective Search proposal algorithm and analyzed its four topmost error modes. We designed the VH-Connect algorithm to compensate two of the four error modes, which even are a problem in many other current proposal algorithms. For the other two error modes we use the Edge Boxes method, which is based on different principles than Selective Search and thus generates somewhat different object hypotheses. This is beneficial for our diversification strategy. We consider the combination of segmentations and edge distribution analysis a rich image representation and therefore a good combination for generating generic object proposals.


\bibliographystyle{IEEEbib}
\bibliography{icme2016template}

\begin{thebibliography}{10}

\bibitem{Wang14}
X.~Wang, M.~Yang, S.~Zhu, and Y.~Lin,
\newblock ``Regionlets for generic object detection,''
\newblock {\em IEEE Transactions on Pattern Analysis and Machine Intelligence},
  2014.

\bibitem{Girshick14}
R.~Girshick, J.~Donahue, T.~Darrell, and J.~Malik,
\newblock ``Rich feature hierarchies for accurate object detection and semantic
  segmentation,''
\newblock {\em IEEE Conference on Computer Vision and Pattern Recognition},
  2014.

\bibitem{Szegedy14}
C.~Szegedy, S.~Reed, D.~Erhan, and D.~Anguelov,
\newblock ``Scalable, highquality object detection,''
\newblock {\em arXiv:1412.1441}, 2014.

\bibitem{Uijlings13}
J.R.R. Uijlings, K.E.A. van~de Sande, T.~Gevers, and A.W.M. Smeulders,
\newblock ``Selective search for object recognition,''
\newblock {\em International Journal of Computer Vision}, 2013.

\bibitem{Zitnick14}
C.~L. Zitnick and P.~Doll{\'a}r,
\newblock ``Edge boxes: Locating object proposals from edges,''
\newblock {\em European Conference on Computer Vision}, 2014.

\bibitem{Alexe10}
B.~Alexe, T.~Deselaers, and V.~Ferrari,
\newblock ``What is an object?,''
\newblock {\em IEEE Conference on Computer Vision and Pattern Recognition},
  2010.

\bibitem{Hosang15}
J.~H. Hosang, R.~Benenson, P.~Doll{\'{a}}r, and B.~Schiele,
\newblock ``What makes for effective detection proposals?,''
\newblock {\em arXiv:1502.05082}, 2015.

\bibitem{Zhu15}
H.~Zhu, S.~Lu, J.~Cai, and Q.~Lee,
\newblock ``Diagnosing state-of-the-art object proposal methods,''
\newblock {\em arXiv:1507.04512}, 2015.

\bibitem{Chen15}
X.~Chen, H.~Ma, X.~Wang, and Z.~Zhao,
\newblock ``Improving object proposals with multi-thresholding straddling
  expansion,''
\newblock {\em IEEE Conference on Computer Vision and Pattern Recognition},
  2015.

\bibitem{Girshick15}
R.~Girshick,
\newblock ``Fast r-cnn,''
\newblock {\em arXiv:1504.08083}, 2015.

\bibitem{Han14}
L.~Han and A.~Han,
\newblock ``An edge detection algorithm based on morphological gradient and the
  otsu method,''
\newblock {\em IEEE International Conference on Bio-Inspired Computing -
  Theories and Applications}, 2014.

\bibitem{Otsu79}
N.~Otsu,
\newblock ``A threshold selection method from gray-level histograms,''
\newblock {\em IEEE Transactions on Systems, Man and Cybernetics}, 1979.

\bibitem{Lin14}
T.~Lin, M.~Maire, S.~Belongie, L.~Bourdev, R.~Girshick, J.~Hays, P.~Perona,
  D.~Ramanan, C.~L. Zitnick, and P.~Doll{\'{a}}r,
\newblock ``Microsoft coco: Common objects in context,''
\newblock {\em arXiv:1405.0312}, 2014.

\bibitem{Deng09}
J.~Deng, W.~Dong, R.~Socher, L.-J. Li, K.~Li, and L.~Fei-Fei,
\newblock ``Imagenet: A large-scale hierarchical image database,''
\newblock {\em IEEE Conference on Computer Vision and Pattern Recognition},
  2009.

\bibitem{Joly09}
A.~Joly and O.~Buisson,
\newblock ``Logo retrieval with a contrario visual query expansion,''
\newblock {\em ACM International Conference on Multimedia}, 2009.

\bibitem{Romberg11}
S.~Romberg, L.~G. Pueyo, R.~Lienhart, and R.~van Zwol,
\newblock ``Scalable logo recognition in real-world images,''
\newblock {\em ACM International Conference on Multimedia Retrieval}, 2011.

\end{thebibliography}

\end{document}